# Coordinated Use of Large Language Models in Academic Research[1]

Vladislav Marinov
Assoc. Prof. Dr., Department of Modern Bulgarian Language
St. Cyril and St. Methodius University of Veliko Tarnovo, Bulgaria
w.marinov@ts.uni-vt.bg · ORCID: 0000-0003-4477-5783

**Abstract**

This study proposes a methodological framework for the coordinated use of multiple large language models in academic research. Its central argument is that involving more models does not in itself guarantee more reliable results; what matters is how cognitive and verification operations are organised. The framework is based on seven principles (procedural independence, functional allocation, the diagnostic value of divergence, the diagnostic value of convergence, external verification, proportionality of checking, and human epistemic responsibility) and sets out a non-linear multi-LLM protocol in which independently obtained results are compared, critically examined and, where necessary, verified against external sources. The approach is illustrated through real-world cases drawn from research practice. The study concludes that moving from a single language model to a coordinated multi-LLM process represents a change in the organisation of research, in which the researcher makes the final decisions and bears scholarly responsibility.

## 1. Introduction

The use of large language models (LLMs) has a place in academic research, ranging from systematising the literature and assisting with the formulation of hypotheses to editing and critically discussing scholarly arguments. In practice, a researcher typically uses one of the available LLMs. Developments in prompt engineering offer opportunities for more precise control over this interaction through carefully formulated instructions, structured context and various strategies for eliciting information from the model. Yet epistemic risks persist even when additional settings are used with a particular language model, such as enabling memory across conversations or creating projects. Common problems include the generation of convincing but factually incorrect statements (so-called hallucinations), changes in responses resulting from variations in the wording of the same task, susceptibility to agreement with the user's position (sycophancy), and discrepancies between potentially available knowledge (encoded in the model during training) and the knowledge actually expressed in a particular response.

---

[1] This is an English translation, prepared by the author, of an article originally published in Bulgarian: Marinov, V. (2026). ‘Координирано използване на големи езикови модели в академичната работа’. *Media and Language*, (00), pp. 1–28. https://doi.org/10.58894/dvzte852.

The argument and conclusions remain unchanged. Editorial adjustments include terminology harmonisation; the English title (the original publication used the English title ‘Coordinated Integration of Large Language Models in Academic Work’); one keyword (‘data verification’ is rendered as ‘verification of claims’); minor corrections to bibliographic details and citation formatting; and the redrawing of Figure 1 in English to align its five phases and possible transitions with the protocol described in Section 3.2. The original Bulgarian publication remains the version of record.

As a result, an increasing number of researchers are incorporating additional LLMs to carry out a single task. From a methodological perspective, however, what matters is not how many models are used, but how tasks are allocated among them and how their interaction is organised throughout the research process. Practice shows that identical responses from several LLMs do not guarantee correctness, since different models may reproduce the same errors owing to overlapping training corpora or shared algorithmic dependencies. Equally, when models generate different responses, there is no indicator of which response is correct.

The present study forms part of a broader investigation into the use of LLMs in academic research and in the teaching of Bulgarian language and literature. Its focus is on how a human researcher can organise their own academic work with the assistance of several different LLMs. Multi-agent systems such as AutoGen, CAMEL, MetaGPT and ChatDev, in which different LLM agents interact to perform a common task, fall outside the scope of this study. When working with several language models, the researcher has a specific role: rather than merely consuming their outputs, the researcher organises and controls the entire process of interaction with the models. This involves determining its structure, controlling the flow of information, identifying when verification is needed, and establishing the grounds for accepting a particular result. This form of activity can be related both to the division of epistemic labour and to the concept of extended cognition. Nevertheless, scholarly responsibility ultimately rests with the human researcher, who defines the tasks, assesses the grounds for convergence and divergence, and bears responsibility for the final conclusion.

The aim of this study is to propose a methodological framework for the coordinated use of multiple LLMs in academic research. Rather than seeking to demonstrate the superiority of any particular language model, it aims to outline a procedure for diagnosing errors and conducting systematic external verification. The underlying assumption is that the methodologically structured use of different LLMs can reveal certain problems that do not emerge, or are not recognised as problems by the user, when interacting with a single model. The advantage of using more than one large language model is procedural: it creates additional points of control and opportunities for verification, provided that each model performs its initial assessment independently, without access to the responses of the others.

The proposed approach is based on seven fundamental principles governing the organisation of interactions between models, the verification of results and the role of the human researcher: (1) procedural independence; (2) functional allocation of operations; (3) the diagnostic value of divergence; (4) the diagnostic value of convergence; (5) external verification; (6) proportionality of checking; and (7) human epistemic responsibility. On this basis, the study proposes a multi-LLM research protocol, whose application is illustrated through specific cases drawn from research practice, followed by a critical analysis of the risks, limitations and possibilities for subsequent empirical testing.

## 2. Theoretical Framework

### 2.1. Limitations of Using a Single LLM and the Need for Checking

A fundamental problem with using a single LLM is that a convincingly formulated response does not guarantee factual reliability. Among the characteristic manifestations of this limitation are so-called hallucinations, cases in which a model generates content presented as relevant to the task but unsupported by the necessary factual information. This risk is particularly pronounced in academic research, where many LLM-assisted operations involve verifiable claims, including bibliographic information, citations, summaries of studies and interpretations of empirical findings.

The factual reliability of generated content has also been investigated in relation to the Bulgarian language. Berbatova and Salambashev (2023) examine different approaches to evaluating hallucinations and propose a method that does not require pre-existing reference data, which they test experimentally on Bulgarian-language material. Their results indicate that the reliability of generated content should be assessed independently rather than inferred from the persuasiveness and internal coherence of a model's response. From a methodological perspective, it is important to distinguish between how a response is formulated and its factual reliability. A confident tone, coherent exposition and appropriate scholarly terminology may create an impression of credibility, but do not in themselves establish whether the information presented is true. The risk is particularly high when users lack sufficient prior knowledge to recognise errors. Hwang and Jeong (2025) experimentally examine whether advance warnings about possible hallucinations reduce users' tendency to accept false information generated by AI. They conclude that such warnings alone are insufficient to prevent the acceptance of misinformation.

Adding a second or third LLM does not provide factual verification either, since different models may reproduce identical or similar errors. Checking by another LLM therefore serves primarily a diagnostic function: it can reveal discrepancies, questionable claims or the need for further verification, but cannot independently establish which claim is true. Where information is factually verifiable, the final verification must be carried out against a reliable external source.

In their analysis of large language models' sensitivity to the wording of instructions (prompt sensitivity), Chatterjee et al. (2024) introduce the POSIX index and find that different types of prompt modification affect responses differently: changes to the prompt template have the greatest impact on multiple-choice tasks, whereas paraphrasing has the greatest impact on open-ended tasks. While conventional evaluations report averaged performance measures, Zhuo et al. (2024) examine prompt sensitivity at the level of individual instances. Their results show that even when average performance remains stable, minor changes in wording can substantially alter the response to a particular question. This directly affects academic research, where the reliability of individual results is essential. Mahaut et al. (2024) additionally investigate how this sensitivity affects the factual confidence of large language models. They find that confidence estimates vary even across semantically identical inputs, calling into question the stability of LLMs' parametric knowledge and demonstrating that variation in wording also affects the persuasiveness of generated responses. This problem is exacerbated during extended conversations with a single language model. The accumulated context establishes a shared frame of reference, making it difficult to distinguish an independent assessment from one influenced by earlier exchanges.

A language model's response may also be influenced by the position of the person asking the question. This risk is associated with sycophancy, the tendency of LLMs to agree with users or affirm their expressed positions. Cheng, M. et al. (2025) examine the issue more broadly through the concept of social sycophancy: the tendency of AI not merely to affirm specific user statements but systematically to endorse users' actions, worldviews and self-assessments. The authors find such behaviour to be widespread among the LLMs examined and, in two experiments involving a total of 1,604 participants, demonstrate its effects on users. Sycophantic responses are rated as higher quality and generate greater trust in the model, while also increasing users' confidence that their own positions are correct.

In academic research, however, this creates a particular risk of mutual reinforcement of the researcher's initial assumption. If researchers include their own hypothesis or preferred interpretation in an instruction, the model may confirm it, and that agreement may subsequently be treated as additional grounds for regarding the hypothesis as correct. The risk may increase during extended conversations, as users' previous positions, assessments and preferences gradually accumulate in the context and may influence the LLM's responses. For this reason, in the proposed multi-LLM approach, the initial assessment should be conducted without first providing the models with the researcher's position or the responses of other models. Only after independent assessments have been obtained should they be compared and used in subsequent checking operations.

Another essential distinction in interpreting LLM responses is that between parametric knowledge, encoded during training, and contextual knowledge, consisting of information supplied in the context of the task. In another study by the author, it was concluded that the presence of parametric knowledge does not guarantee that it will be used in a particular model response (Marinov, unpublished manuscript). These findings are consistent with earlier observations by Cheng, S. et al. (2025), who show that even when particular knowledge is encoded in an LLM's parameters, this does not guarantee its use when generating a response.

The authors investigate this relationship using their ECHOQA benchmark, which comprises tasks designed to trace the interaction between parametric and contextual knowledge. They first establish whether particular information is accessible to the model and then examine whether the same information is used after different types of contextual knowledge have been introduced. Their findings indicate that the activation of parametric knowledge can be blocked by context, regardless of whether that context supplements or contradicts the available information. When responding to questions requiring specific scientific knowledge, the models examined fail to use previously established parametric knowledge in more than 60% of cases. The authors also demonstrate that specially formulated instructions can improve the use of available parametric knowledge, confirming that the generated response depends on the particular conditions under which the model performs the task.

Zhao et al. (2025) investigate a different aspect of the same problem. Using a method for tracing information about specific entities through different model layers, they analyse how parametric and contextual knowledge are processed when the two conflict. Their results show that information supplied in the context and knowledge already acquired during training are processed differently. Even when they contradict one another, both types of information may remain present during processing. This indicates that the resulting response is not simply a reproduction of one source of knowledge or the other, but arises from their interaction. The absence of particular knowledge from a specific response does not necessarily mean that the model lacks that knowledge in its parametric memory.

The four groups of limitations discussed above – (1) hallucinations, (2) the influence of instruction wording on responses, (3) sycophancy and (4) instability in the use of available parametric knowledge – show that introducing an additional model to perform a checking function may help identify errors or problems in the initial result, but does not guarantee greater reliability. Comparing independently generated model outputs can reveal two distinct diagnostic signals that may guide the researcher towards further verification. On the one hand, agreement between independently obtained results indicates that a particular result is not specific to a single LLM, but does not in itself establish its factual accuracy. On the other hand, when models produce different responses, there is no marker indicating which result is correct. In this sense, the coordinated multi-LLM approach shifts the focus from selecting a particular LLM to the function of each individual operation, the conditions under which it is performed, and how its result is interpreted and used in subsequent stages of the research.

### 2.2. The Use of Multiple LLMs in the Context of Contemporary Research

With the development of large language models, increasing attention has been paid to methods for formulating effective instructions (prompt engineering), which allow the capabilities of a model to be used more fully without modifying its parameters (Sahoo et al., 2024; Schulhoff et al., 2024). In academic settings, this enables more precise control through structuring the context, requesting justification and progressively reformulating the task. However, a fundamental limitation must also be recognised: even minor changes in the wording of a task, without altering its meaning, can produce different model responses (Chatterjee et al., 2024; Salinas and Morstatter, 2024). Because of this dependence, prompt optimisation cannot be regarded as a substitute for independent checking of the result. In the former case, the conditions under which the model generates a response are modified; in the latter, an already generated claim is subjected to an assessment that does not depend directly on the same context. In other words, optimising instructions is intended to obtain a more appropriate response from a particular model, whereas organising work with several models involves successively assessing and critically checking the results obtained.

This approach must also be distinguished from the use of multi-agent systems, in which complex tasks are typically decomposed among agents performing separate roles through different LLM instructions. The MAS-ProVe study (Venkataramani et al., 2026) notes that increasing the number of participating agents creates more opportunities for potential errors, while automated verification of individual stages does not necessarily guarantee a better final result. At the same time, there is evidence that multiple checking operations can improve results even when the same underlying LLM is used both to generate and to check a response (Lifshitz et al., 2025). The authors also allow checking agents to be based on different LLMs, with diversity achieved by varying the underlying model, the aspect being checked and the checking strategy. The HLER approach (Zhu and Wang, 2026), in which specialised agents perform different tasks within the research process, is also illustrative. Here, too, the authors identify the shared underlying model as a limitation: the agent reviewing the resulting manuscript uses the same underlying LLM as the agent that produces it. Although the two agents have different roles and instructions, using the same LLM for writing and reviewing the text constitutes a limitation and reduces the credibility of the assessment. Involving another LLM or human reviewers, by contrast, would allow the quality of the resulting text to be checked by an independent source. Thus, the functional division of tasks among several agents operating within a single language model and the use of different LLMs are two distinct characteristics of the research process.

Tsukada's Observer-Mediated Multi-LLM Triangulation (OMT) method comes closer to the problem addressed here, since it introduces independent queries to different models and subsequently assigns a human observer the task of classifying discrepancies between their responses (Tsukada, 2026). OMT deliberately excludes communication and mutual criticism between models in order to preserve the independence of their initial assessments. However, the limitations of this procedure are also identified: when different models make identical or similar errors, agreement between their responses may create a misleading impression of correctness, and the procedure may not be triggered.

The discussion so far indicates that one of the central questions arising from the inclusion of LLMs in the research process concerns the trustworthiness of model responses, that is, the extent to which researchers can trust them. Ferrario, Facchini and Termine (2024) examine whether high AI accuracy is sufficient grounds for attributing expertise and epistemic authority to such systems. They distinguish between the two concepts, associating expertise with specialised knowledge and abilities, and epistemic authority with trust in the judgement of a source possessing greater competence in a particular field. From this perspective, they challenge the possibility of attributing full epistemic expertise to AI, since this requires not only the ability to provide correct answers but also understanding and certain intellectual capacities that AI systems do not currently possess. They allow AI systems only partial expertise, associated with their capacity to store and process large amounts of information and, in most cases, to generate correct statements. At the same time, the authors develop the idea that successful human–AI interaction may give rise to a hybrid epistemic subject to which expertise and authority can, under certain conditions, be attributed.

Hauswald (2025) advances a different position. He examines the issue in relation to the classical model of testimonial authority, in which a person accepts a claim because it has been communicated by someone with greater competence in the relevant field. Such a relationship presupposes that the source believes the claim and communicates it consciously. Since a language model has no subjective understanding and does not communicate consciously, it does not meet these criteria and cannot function as an authentic epistemic witness. Hauswald nevertheless argues that AI can perform a functionally similar role by providing information and supporting understanding in fields where users have substantially less competence. He therefore allows for an extension of the concept of epistemic authority: in the case of AI, trust is grounded not in testimony in the classical sense, but in reliable indicators that a particular system produces correct results in the relevant domain.

A similar argument is developed by Majszak and Sartori (Majszak and Sartori, forthcoming). They argue that differences between humans and machines are not, in themselves, sufficient grounds for rejecting the possibility that AI can function as an epistemic authority. In their view, the decisive question is whether a system is sufficiently reliable in the relevant field to provide grounds for trusting its results, much as one would trust a human expert. In this sense, they regard AI as a candidate for epistemic authority.

The question of epistemic responsibility in the use of language models is also addressed by Stoyanov (2026), who proposes an ethical framework for evaluating individual claims obtained with the assistance of an LLM. The main parameters are the degree of verification and the assumption of responsibility by a human subject. However, this framework concerns interaction between a human and a single LLM and does not address the specific problems arising from comparing and coordinating results from several different language models. When more than one LLM is used, the task becomes more complex: the researcher must compare potentially divergent results and decide which source to trust. In this connection, Hauswald (2025) discusses the so-called user/2-AI problem, by analogy with the novice/2-expert problem, in which a less competent person must choose between conflicting opinions offered by two experts. The situation with language models is analogous: two apparently reliable systems may provide different answers, leaving the user to determine how to resolve the conflict. According to Hauswald, this requires further assessment of the reliability of both sources.

Majszak and Sartori examine a related problem through the concept of peer disagreement with a machine, a situation in which a human and an AI arrive at different answers to the same question (Majszak and Sartori, forthcoming). In this case, the machine may be regarded as a source of comparable competence on the basis of its previous performance (track record), that is, the consistency with which it has produced correct results. Consequently, when disagreement occurs, the decision should not automatically favour either the human or the AI, but should take account of the grounds for trusting the particular source and its reliability in the relevant domain.

In the framework proposed here, however, different LLMs are not treated as a collective of authorities whose agreement automatically increases the probability that a claim is true, but as sources of results serving different diagnostic functions. The human researcher determines the sequence of operations, evaluates the arguments and bears responsibility for incorporating the results into scholarly discourse. In other words, the human role is not diminished by distributing individual cognitive operations among models; on the contrary, distinguishing their functions and evidential status makes that role more explicit.

A different perspective on this problem is offered by the concept of extended cognition, associated with the idea of active externalism, according to which external resources may, under certain conditions, participate as constituents of the cognitive process itself. Clark and Chalmers (1998) challenge the restriction of cognition to processes occurring within the human brain and allow for the formation of a coupled system comprising a human and an external resource used by that person. Applying this perspective to LLMs, however, requires qualification: using an external resource in a cognitive process is not sufficient for that resource to be regarded as part of an extended cognitive system. This requires closer functional integration, in which the external resource does not merely provide information but participates systematically in the cognitive process itself.

A similar understanding of the role of technology in cognitive activity is developed by A. L. Semenov and K. E. Ziskin (Semenov and Ziskin, 2022). Drawing on L. S. Vygotsky's instrumental approach, the authors regard digital technologies and artificial intelligence as tools of human intellectual activity whose incorporation does not merely add new capabilities but may change how the activity itself is carried out. This perspective is important for the present study because it allows the use of LLMs in academic research to be understood not only as the use of an additional tool, but also as an opportunity to organise individual cognitive operations differently.

It is precisely this transformation of cognitive processes through contemporary AI systems that is the focus of the study by P. R. Smart, R. W. Clowes and A. Clark, who analyse systems using retrieval-augmented generation (RAG), including the Digital Andy model (Smart, Clowes and Clark, 2025). In RAG, a language model does not rely solely on information acquired during training; when performing a particular task, it also receives additional information retrieved from an external source by another component of the system. The authors show that merely adding such a resource to an LLM is insufficient to establish functional integration in the strong sense associated with extended cognition. Particularly revealing is the question of who initiates the relevant operation: in Digital Andy, retrieval is not initiated autonomously by the LLM but by an application external to it. In this sense, the LLM is a passive recipient of the information provided rather than the initiator of its retrieval. This distinction matters for the present framework because the coordinated multi-LLM process likewise does not presuppose an autonomous cognitive system. The researcher explicitly initiates independent assessment, criticism, comparison or external verification and determines when the result of one operation should provide grounds for the next.

On this basis, the coordinated multi-LLM research process can be defined as a way of organising academic work in which different large language models are involved at separate stages of research to perform specific cognitive and checking functions, without being assigned permanent roles. It is also important to emphasise that the researcher manages the sequence of operations and the exchange of information between models, makes the final decisions and bears scholarly responsibility.

Three interconnected dimensions of the process follow from this definition. At the model level, using more than one LLM reduces dependence on a single language model without presupposing complete statistical independence between the models involved. At the functional level, individual systems may perform different cognitive operations according to the situation – generation, independent assessment, criticism or checking – without these functions becoming their permanent roles. At the methodological level, the researcher organises the relationships between these operations, controls the exchange of information and determines when model-based checking should give way to external evidential grounds.

## 3. Methodological Framework

The proposed methodological framework brings together the theoretical perspectives and limitations discussed above into a system of principles and procedures for the coordinated use of multiple LLMs. It does not prescribe a universal sequence of actions to be applied uniformly to every academic task. Instead, it specifies the conditions under which different LLMs are involved at particular stages of the research process according to (1) the function they perform, (2) the information to which they have access, and (3) how the resulting output should be assessed and checked. In other words, the organisation of the process is determined not by the choice of a particular model, but by the function to be performed at a given stage of the research and the conditions under which an LLM can be used to perform it.

### 3.1. Principles of the Coordinated Multi-LLM Process

The framework rests on seven interconnected principles. These are not sequential stages, nor do they imply that every task must activate each principle to the same extent. The first two establish the conditions under which model operations are organised and differentiated. The next two define the diagnostic status of relationships between independently obtained results. The fifth establishes the boundary between model-based assessment and external evidential grounds. The sixth regulates the intensity of checking, while the seventh identifies the subject responsible for the final epistemic judgement. These principles are not intended to produce agreement between models automatically. Rather, they organise the process so that differences and uncertainties in the results can be identified and subjected to further checking.

**1. Procedural independence.** Independence is a condition of a particular operation rather than an inherent property of the models used. Different LLMs may have overlapping training data, similar architectural solutions and shared dependencies, so complete statistical or epistemic independence between them cannot be assumed. What the researcher can control is the information environment in which each assessment takes place. When the aim is to obtain an independent model assessment, the second model should not be given the first model's response, its reasoning or the researcher's assessment in advance. After the initial independent assessment, this separation may be deliberately discontinued. One model's response may be provided to another for criticism, examination of the reasoning or identification of omissions. This operation is no longer an independent assessment but a dependent critical checking operation, and must be interpreted accordingly.

**2. Functional allocation of operations.** In the coordinated process, functions are not permanently assigned to particular models. The framework does not assume that one LLM is a 'generator', another a 'critic' and a third a 'verifier'. Such fixed assignments would turn the contingent characteristics of the selected systems into a component of the methodology and limit the framework's reproducibility when the available models change. Each model's function is determined by the particular operation. The same model may, at different points, generate an initial hypothesis, independently assess another hypothesis, criticise a supplied argument or analyse a discrepancy between two positions. This allows the methodological description to remain independent of any particular technological configuration.

Different methods may be used to check results in a coordinated multi-LLM process. In independent model assessment, another LLM performs the same task without access to the response already obtained. In cross-model critical checking, the result or reasoning produced by one model is provided to another, which is asked to analyse it and identify possible errors, weaknesses or contradictions. Checking may also proceed to a further level: another model may assess not the original response but the criticism or checking already performed. This operation is termed meta-checking. External verification differs from the preceding operations because it establishes the credibility of a claim by comparing it with primary data, a regulatory document, an original publication or another reliable source outside the LLMs used.

The limitations of model-based assessment have also been established empirically in research involving Russian. Pugachev, Fenogenova, Mikhailov and Artemova (2025) compare human assessments of model-generated responses with assessments produced by eight LLMs. Their analysis reveals incomplete agreement between human and model assessments, with models showing uneven ability to recognise different types of errors. These findings support the distinction between checking by another LLM and checking against an objective criterion. Introducing a model in an evaluative role may reveal problems in the initial result, but its assessment must itself be critically examined. The methodological value of such checking therefore lies less in determining the correct result than in identifying discrepancies that require further verification.

**3. Diagnostic value of divergence.** Divergence between independently obtained results provides grounds for identifying the cause of the discrepancy and, where necessary, undertaking further checking. It shows that different models may arrive at different results for the same task, thereby revealing uncertainty that might remain unnoticed when only one LLM is used. Discrepancies between responses may arise from different interpretations of the task, differences in how models use available knowledge and construct their reasoning, context-specific influences, or factual errors made by one or both systems. The appropriate response depends on the cause. Different interpretations of the task may require clarification of the instructions and reassessment; differences in the knowledge used or the reasoning may call for cross-model criticism; contextual influences may require checking under altered conditions; and factual contradictions require verification against external evidence.

**4. Diagnostic value of convergence.** Agreement between independently obtained results also has diagnostic value, but should not be interpreted as proof of truth. As already noted, different LLMs may reproduce the same error because of similar training data, shared cultural or discursive dependencies, or similar generation strategies. Such agreement indicates that the result is not specific to a single system and provides grounds for examining the relevant hypothesis further, but does not independently confirm its credibility. Diagnostic convergence therefore serves as a guide to the direction of research rather than a substitute for the researcher's further external verification of facts.

**5. External verification and evidential adequacy.** Where claims can be verified empirically or against documentary sources, checking should not be confined to mutual assessment between different LLMs. A third model may analyse the second model's criticism, and a fourth may evaluate that analysis, but adding further systems does not change the nature of the procedure: it remains model-based checking. If a claim can be checked against primary data, an original publication, a regulatory document or another reliable source, this should take precedence over involving additional models.

External verification involves more than identifying a thematically relevant source. It is necessary to establish what the source actually states and the extent to which it supports the claim for which it is cited. The procedure therefore comprises at least three distinguishable operations:

*Identification of a relevant source → checking its actual content → assessment of its evidential adequacy in relation to the specific claim.*

This distinction is particularly important when working with scholarly literature. A study may be thematically related to the claim being made without its findings constituting direct evidence for that claim. Similarly, two studies may reach similar conclusions at different analytical levels without providing two independent confirmations of the same hypothesis. External verification is therefore not a final formal step following model comparison, but a separate evidential operation.

**6. Proportionality of checking.** Not every model-generated claim requires the same intensity of checking. Attempting to check every sentence using multiple models and external sources would make the procedure practically unworkable. The degree of checking should be proportionate to epistemic risk: the likelihood that an error will remain undetected and the consequences of incorporating it into a scholarly result.

Three indicative levels of the need for checking can be distinguished within the framework. A low need for checking mainly characterises editorial or formal operations in which the model introduces no new factual or interpretative content. A moderate need for checking arises in synthesis, interpretation or argument formulation, where the result requires critical assessment but not every claim requires immediate external verification. A high need for checking applies to factual claims, bibliographic information, quotations, regulatory provisions, empirical findings and other elements whose inaccuracy would affect the evidential basis of the research.

These levels do not constitute a measurement scale and are not intended as a universal classification. They provide guidance for procedural decisions: the greater the risk of an epistemically significant error, the stronger the grounds for independent and/or external checking. Proportionality also determines when checking should stop. An additional model operation is justified only if it can address a specific unresolved question. When the next step would merely repeat information already obtained without reducing the relevant uncertainty, continuing the sequence of model operations has no methodological value.

**Table 1.** Levels of checking and corresponding procedures

| Level | Types of tasks | Procedure |
|---|---|---|
| **Low** | Heuristic tasks, suggestions for structure, language editing | Single model operation + human judgement |
| **Moderate** | Terminological decisions, interpretation of an argument | Independent assessment, criticism or comparison with the literature |
| **High** | Empirical data from other studies, bibliographic information, numerical values, quotations | Mandatory external verification against a primary source |

**7. Human epistemic responsibility.** The coordinated multi-LLM process distributes cognitive operations among different LLMs, but the researcher bears ultimate scholarly responsibility. The researcher defines the task, decides when an independent assessment is needed, determines how convergence or divergence should be interpreted, identifies the relevant external grounds and bears responsibility for incorporating the result into the scholarly text.

This principle is also essential in automated or semi-automated configurations. The fact that one system can generate a response, another can criticise it and a third can assess the criticism does not automatically create a reliable epistemic chain. Human oversight does not consist merely in subjectively selecting the response that the researcher considers best, but in determining the evidential status of each operation. At the same time, the human role should not be idealised as a neutral corrective to model limitations. Researchers may terminate checking without sufficient grounds when the result agrees with their initial hypothesis, appears subjectively convincing enough or is supported by agreement between models.

### 3.2. Operationalising the Principles through a Multi-LLM Research Protocol

The seven principles can be operationalised through a protocol that organises the sequence of model-based and non-model operations. The protocol is not a fixed algorithm that every task must follow in full. It is non-linear: transitions between steps are dynamic and depend directly on the outcome of the preceding operation. Accordingly, the number of LLMs used is not a predetermined criterion for a 'multi-LLM' process, but follows from the particular structure of checking.

The protocol comprises five phases.

**Phase 1. Defining the task and epistemic risk.** The researcher identifies the type of expected result and the level of epistemic risk in order to determine whether checking is necessary and, if so, what kind.

**Phase 2. Initial model operation.** The selected model performs a precisely defined function, and its output is treated as an initial hypothesis rather than established knowledge.

**Phase 3. Independent assessment and diagnosis.** Another model performs the same task without access to the first result, and comparison of the two responses identifies convergence or divergence.

**Phase 4. Controlled critical checking and external grounds.** The separation between models is deliberately ended by the researcher to allow cross-model criticism or meta-checking, while verifiable claims are checked against an external source.

**Phase 5. Synthesis, decision and documentation.** The researcher synthesises the results, makes the final decision and documents the process in proportion to its significance.

The five phases do not form a mandatory linear sequence. Phase 2 may lead directly to external verification if a claim is factually verifiable. After Phase 3, no further model operation may be necessary. Phase 4 may lead back to a new independent assessment if the criticism has changed the formulation of the problem itself. What matters is not the number of phases completed, but the methodological necessity of each one.

The general rule is that a new model operation is initiated only when a specific unresolved question has been formulated for which that operation can provide relevant diagnostic information. The protocol thus prescribes not a predetermined number of model operations, but a rule governing when they should be initiated and terminated.

The phases described above and the possible transitions between them are summarised in Figure 1.

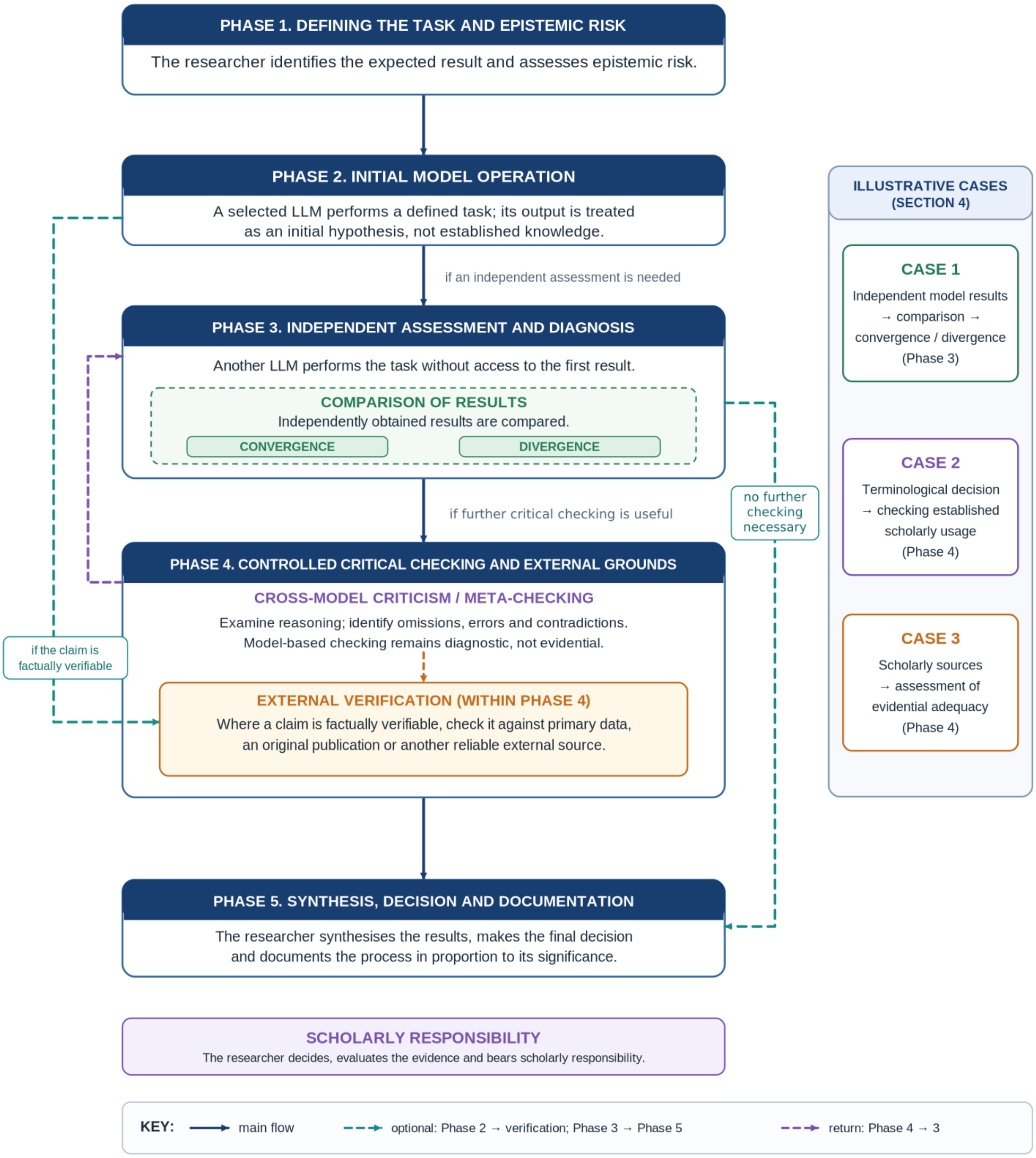

**Figure 1.** Main operations and possible transitions in the coordinated multi-LLM research process

## 4. Application of the Protocol: Reconstruction of Real Research Cases

To demonstrate how the proposed framework can be applied to real academic tasks, three cases drawn from research practice are examined. They were selected to represent different types of problems identified in the theoretical part of the study and the corresponding methods of checking.

The first case concerns an earlier comparative study of four LLMs using tasks in the field of Bulgarian language, in which the research design incorporated conditions for independent task performance and restrictions on information exchange between models from the outset (Marinov, unpublished). This case makes it possible to examine the diagnostic function of convergence and divergence when independently obtained model results are compared. The second case illustrates a different type of problem: the selection of a term whose appropriateness must be checked against established usage in the relevant academic field. The third case demonstrates the need to distinguish the thematic relevance of a publication from the evidential value of its specific findings. The three cases thus cover different situations in which the outcome of one operation determines the need for, and nature of, the next checking operation: from comparing independently obtained model results, through checking a selected term, to assessing the evidential value of a scholarly source. They illustrate different functions of the proposed protocol in actual research practice.

In the first case, ChatGPT, Gemini, Claude and BgGPT were given an identical diagnostic test in Bulgarian language, comprising 34 tasks in phonetics, morphology, syntax, and lexicology and phraseology. The models completed the test in separate sessions with memory disabled and without access to the responses of the other models. The entire test was provided at once, and follow-up questions were asked only when responses were incomplete or internally contradictory (Marinov, unpublished). From the perspective of the present framework, this design largely satisfies those operational conditions for procedural independence that the researcher can control.

Comparative evaluations of language models on Bulgarian-language tasks have also been conducted by Hardalov et al. (2023), who propose bgGLUE, a comprehensive benchmark for evaluating Bulgarian language understanding, and identify differences in performance according to task type. It should be noted, however, that the models examined belong to an earlier generation of encoder-based architectures, rather than the generative LLMs considered here. The more recent Bulgarian Massive Multitask Language Understanding benchmark (Koeva et al., 2026) extends comparative evaluation to tasks from different subject areas and likewise identifies differences in the performance of the LLMs examined across domains.

In the present study, however, comparing the results serves a different purpose: it is used not only to identify differences between models but also to determine whether further checking is needed and, if so, in what direction it should proceed. Comparison of the four independently obtained assessments reveals both convergent and divergent errors. Convergence on some tasks, particularly at the phonetic level, suggests that the difficulties may not be specific to individual models, whereas divergence on other tasks reveals differences in the consistency and nature of the errors made. In both cases, comparison serves a diagnostic rather than an evidential function. Similar results may point to a more general problem, while differences may indicate the need for further checking, but the distribution of responses across models does not itself establish the credibility of the result.

The procedure can be summarised as follows:

*Independent assessments → convergence/divergence → identification of the problem → external expert verification.*

Similar instability has also been observed in studies examining LLM performance in relation to Bulgarian orthographic norms. In an experiment involving ChatGPT, Gemini and Copilot, A. Todoranova (2024) examines model performance across three successive stages: initial completion of the task, provision of the relevant orthographic rules, and introduction of misleading statements. The results show improvement after the rules are provided, but also subsequent changes in some responses under the influence of false contextual information. Another study by Todoranova and Todoranova (2026) demonstrates the same tendency: after being provided with the official language rule, the models examined achieve high or complete accuracy, but their performance deteriorates following the subsequent introduction of misleading contextual information. The observed similarity in the behaviour of different models under changing contextual conditions is also relevant to the present study. Agreement between model results may indicate their stability under particular conditions, but is not, in itself, sufficient grounds for assessing their credibility.

The second case concerns checking a terminological decision associated with one of the principles of the present framework. The term convergent validation was initially selected to describe the procedure in which independently obtained results from different LLMs are compared. Subsequent examination of the methodological literature, however, showed that this term has a specific established meaning, dating back to the classic study Convergent and Discriminant Validation by the Multitrait-Multimethod Matrix by Campbell and Fiske (1959), and does not correspond to the function the concept is intended to perform here. The initially selected term convergent validation was therefore revised, and convergence is understood in the present framework in a diagnostic rather than a validation sense.

This case demonstrates that terms associated with established scholarly terminology must be checked against the primary specialist literature. The sequence of operations can be represented as follows:

*Selection of a term → emergence of doubt → external methodological verification → reformulation.*

In other words, the outcome of an operation is not accepted or rejected mechanically; rather, it determines what needs to be clarified and how checking should proceed.

The third case concerns checking thematically relevant scholarly sources dealing with the interaction between parametric and contextual knowledge in LLMs. Such checking must establish not only whether a study addresses the same problem, but also whether its specific findings actually support the claim for which it is cited.

This distinction is clearly illustrated by the studies of Cheng, S. et al. (2025) and Zhao et al. (2025), discussed above, which address the same general problem but provide different types of evidence. As noted earlier, Cheng, S. et al. (2025) use ECHOQA to establish, at the behavioural level, how the presence of contextual information affects the use of parametric knowledge. Zhao et al. (2025), by contrast, investigate the internal mechanisms underlying the interaction between the two types of knowledge by tracing their representation and reconciliation in the model's hidden layers.

Because the data obtained are of different kinds, the two studies can support claims of different scope. This demonstrates that the thematic similarity of a publication is not sufficient in itself. Consequently, the existence of two thematically relevant sources does not amount to two independent confirmations of the same claim.

The sequence of operations in this case can be represented as follows:

*Relevant source → checking its actual content → assessment of evidential adequacy → narrowing the claim.*

The three cases differ in the particular problems addressed and the methods used to check them, but they demonstrate the same fundamental principle: the outcome of an operation determines the need for, and type of, the next checking operation. In the first case, the research design requires different LLMs to perform tasks independently, without access to the other models' results. In the second, the selection of a term important to the study requires further clarification. In the third, checking a relevant source requires assessment of its evidential adequacy.

The general principle is that the choice of the next operation is determined by the question left unresolved by the preceding one. The cases examined show that adding further LLMs is not a universal solution. They demonstrate the analytical applicability of the proposed framework but do not constitute empirical validation of its effectiveness. The cases do not provide sufficient grounds for concluding that the coordinated use of multiple LLMs produces greater accuracy or reliability than working with a single model. Such a conclusion could be drawn only after a larger-scale controlled comparison of the two approaches.

## 5. Opportunities, Limitations and Research Prospects

The proposed framework shifts the focus from the number of language models used to the organisation of research operations. The cases and theoretical perspectives discussed above show that the multi-LLM approach is not simply an extended version of interaction with a single LLM. The methodologically significant change occurs when generation, independent assessment, critical checking, comparison and recourse to external grounds are distinguished as operations with different functions and evidential status, and when individual models' access to previous results is managed according to the function of the particular checking operation. Using multiple LLMs allows the same task to be examined independently and may reveal differences or problems that would remain unnoticed when working with only one model, but it does not in itself guarantee the quality of the final result.

It may be objected that some of the principles of the proposed framework are already part of established research practices and critical thinking, and that their distinctive character here derives solely from their application to LLMs. This objection is justified to some extent.

Independent assessment, critical checking, the search for counterarguments and verification against external sources are not new methodological principles, and the present study does not present them as such. What is distinctive here is how these principles are applied and combined when working with large language models. Repeating a task with the same model preserves dependence on the accumulated conversational context, whereas assigning it to another model without access to the previous result provides procedural independence. Similarly, agreement between several LLMs does not constitute independent evidence of the credibility of a claim, while disagreement requires the cause of the difference to be identified and an appropriate method of further checking to be selected.

The need for such distinctions is related to characteristic features of LLMs discussed above: sensitivity to task wording, the generation of convincing but factually unsupported claims, a tendency towards excessive agreement with the user's position, and the possibility that available parametric knowledge may not be expressed in a particular response. The framework therefore does not propose a substitute for critical thinking in general, but specifies how it can be applied procedurally when working with LLMs, taking account of the particular ways in which these systems may generate unreliable or incorrect information.

Procedurally structuring critical checking does not, however, eliminate the limitations associated with using LLMs. Different LLMs cannot be regarded as completely independent sources because of possible overlaps in training data, similar architectural and post-training approaches, and shared informational dependencies. As noted above, diagnostic convergence should not be turned into a quantitative criterion of credibility whereby a result is considered more reliable merely because more models have arrived at it. The same applies in the opposite direction: divergence increases the diagnostic information available but does not, in itself, establish that an error has occurred.

Working with multiple LLMs requires more time and effort from the researcher, including the comparison and synthesis of the results obtained. The scope of checking must therefore reflect the significance of the particular task and the risk of error. The principle of proportionality does not imply restricting necessary checks, but avoiding additional operations that would not make a substantial contribution to the reliability of the result. It is important to establish in advance not only the conditions under which further checking is necessary, but also those under which it may be terminated.

The central role of the researcher must be understood as both a condition and a limitation of the framework. The researcher's intervention in the process is not neutral and does not automatically guarantee the elimination of model biases. Increasing the number of available responses also expands the scope for selective choice. A researcher may continue checking when a result contradicts their initial hypothesis but stop at the first convenient agreement; prefer the argument of the model that supports their own position; or apply different standards of scrutiny to confirming and disconfirming evidence. In this way, using multiple LLMs expands the opportunities for mutual checking but simultaneously creates a greater risk that researchers will favour results supporting their prior positions.

For precisely this reason, human epistemic responsibility should not be understood merely as the final right to choose between model-generated suggestions. It includes responsibility for the design of the procedure itself: when models need to work independently without access to one another's results; when mutual influence may be permitted; what type of evidence is relevant; how long checking should continue; and what degree of uncertainty should be retained in the final claim. Procedural traceability thus serves a dual function: it makes dependencies between model operations more visible while also reducing the possibility that human decisions will remain concealed behind an apparently objective 'consensus' between systems.

The present study proposes and theoretically substantiates a procedural framework but does not experimentally test its comparative effectiveness. The cases analysed demonstrate its applicability to different types of academic decisions but do not permit the conclusion that the coordinated multi-LLM process necessarily produces more accurate or reliable results.

The next step would be to conduct a controlled comparison between working with a single LLM, several independent sessions with the same model, several different LLMs without a predefined structured procedure, and a coordinated multi-LLM process following the proposed protocol. Such a design would make it possible to distinguish the effect of model heterogeneity from that of procedural organisation and to establish for which types of tasks the separation of operations genuinely improves error detection, critical assessment of arguments and the management of uncertainty.

Empirical testing should cover not only the accuracy of the final result but also the cost of the procedure: the number of operations required, the time and resources needed to perform them, the proportion of errors detected through independent or cross-model checking, and the degree of residual uncertainty. Future research could thus establish not merely whether using multiple models improves results, but under what conditions, and at what cost in time and resources, the coordinated allocation of model operations provides a genuinely measurable methodological benefit.

## 6. Conclusion

The present study has proposed and theoretically substantiated a methodological framework whose application has been illustrated through three cases drawn from academic practice. Comparing the theoretical perspectives, the principles formulated and the cases examined makes it possible to identify the opportunities and limitations of coordinated work with multiple LLMs. Several main conclusions can be drawn from the analysis:

1) Involving multiple LLMs does not in itself increase the reliability of academic work. What matters is how individual tasks are allocated among them and how access to previously obtained results is managed. Repeating a task with a model that has access to the accumulated conversational context preserves dependence on the previous interaction, whereas assigning the task to another model without access to the previous result provides procedural independence. The information conditions under which each individual task is performed are therefore an essential component of the research design.

2) Convergence and divergence between independently obtained model responses serve primarily a diagnostic function. Agreement may show that different models produce the same result when performing a task independently, but it does not constitute independent evidence of its credibility, since the models may make similar errors. Disagreement, by contrast, indicates uncertainty and may draw attention to differences in facts, assumptions, interpretation or argumentation. Consequently, neither agreement nor disagreement between different LLMs determines in itself which result should be accepted. Rather, both serve to identify the need for, and type of, subsequent checking.

3) Checking by other LLMs has clear limitations. Independent assessment, cross-model criticism and meta-checking may reveal errors, contradictions and dependencies that would remain unnoticed when working with a single model, but they do not replace verification against a source external to the model system. Where a claim can be verified against primary data, an original publication, a regulatory document or another reliable source, such verification takes precedence over involving additional models. The cases analysed also demonstrate the importance of distinguishing between the thematic relevance of a source and the extent to which its findings can actually provide grounds for the claim they are used to support.

4) Allocating individual tasks among several LLMs does not mean transferring scholarly responsibility to the models. The researcher organises the process, decides when independent task performance is necessary, selects the appropriate type of subsequent checking, assesses the reliability of the sources used and determines what degree of uncertainty is acceptable in the final result. At the same time, this central role creates a risk that the researcher's own prior expectations and preferences will influence the process. A clear description of the individual steps and the ability to trace them are therefore necessary not only for monitoring dependencies between model results but also for critically assessing the decisions made by the human researcher.

The main contribution of the proposed framework lies not in the idea of using more language models to solve a particular task, but in systematising the conditions under which their results can be incorporated into an organised research process. Further empirical research may establish for which tasks coordinated work with multiple LLMs offers a measurable advantage over using a single model or working with several models without a structured procedure. Distinguishing the generation of an initial result, independent assessment, critical checking, comparison and verification against external sources allows each operation to be used according to its specific function and evidential value. Thus, moving from working with a single large language model to a coordinated multi-LLM process does not simply mean increasing the number of models. It represents a different way of organising academic work, in which the capabilities of these systems are used for generation, criticism and checking, while the researcher makes the final judgement concerning the scholarly soundness of the results and bears responsibility for them.

## Acknowledgements

This study was carried out as part of the research project “Spoken Language in Digital Format: From Archive to Research Resource” („Разговорната реч в дигитален формат – от архив към научен ресурс“, ФСД-31-354-14/28.04.2026).